\documentclass[letterpaper, 10 pt, conference]{ieeeconf} 

\IEEEoverridecommandlockouts 

\usepackage{placeins}
\usepackage{float}
\usepackage{amsmath} 
\usepackage{graphicx}
\usepackage{changepage}
\usepackage{booktabs}
\usepackage{caption}
\usepackage{comment}
\graphicspath{ {./figures/} }
\usepackage[font=small]{caption}
\usepackage{url}
\usepackage{cite}
\usepackage{subfigure}

\title{\LARGE \bf
FlightAR: AR Flight Assistance Interface with Multiple Video Streams and Object Detection Aimed at Immersive Drone Control
}

\author{Oleg Sautenkov*, Selamawit Asfaw*, Yasheerah Yaqoot* \\ Muhammad Ahsan Mustafa, Aleksey Fedoseev, Daria Trinitatova, and Dzmitry Tsetserukou\\
\thanks{The authors are with the Intelligent Space Robotics Laboratory, Center for Digital Engineering, Skolkovo Institute of Science and Technology. 
{\tt \{oleg.sautenkov, selamawit.asfaw, yasheerah.yaqoot, ahsan.mustafa,  aleksey.fedoseev, daria.trinitatova, d.tsetserukou\}@skoltech.ru}}
\thanks{*These authors contributed equally to this work.}
}

\begin{document}

\maketitle
\thispagestyle{empty}
\pagestyle{empty}

\begin{abstract}


The swift advancement of unmanned aerial vehicle (UAV) technologies necessitates new standards for developing human-drone interaction (HDI) interfaces. Most interfaces for HDI, especially first-person view (FPV) goggles, limit the operator's ability to obtain information from the environment. This paper presents a novel interface, FlightAR, that integrates augmented reality (AR) overlays of UAV first-person view (FPV) and bottom camera feeds with head-mounted display (HMD) to enhance the pilot's situational awareness. Using FlightAR, the system provides pilots not only with a video stream from several UAV cameras simultaneously, but also the ability to observe their surroundings in real time.
User evaluation with NASA-TLX and UEQ surveys showed low physical demand ($\mu=1.8$, $SD = 0.8$) and good performance ($\mu=3.4$, $SD = 0.8$), proving better user assessments in comparison with baseline FPV goggles. Participants also rated the system highly for stimulation ($\mu=2.35$, $SD = 0.9$), novelty ($\mu=2.1$, $SD = 0.9$) and attractiveness ($\mu=1.97$, $SD = 1$), indicating positive user experiences. These results demonstrate the potential of the system to improve UAV piloting experience through enhanced situational awareness and intuitive control. The code is available here: https://github.com/Sautenich/FlightAR
\end{abstract}

\section{Introduction}
Unmanned aerial vehicles (UAVs) are quickly emerging as essential tools in a variety of areas, such as agriculture \cite{menshchikov2021real}, logistics\cite{kalinov2020warevision}, seismic exploration\cite{yashin2022autonomous}, etc. 
Also, UAVs are increasingly utilized across diverse operational contexts, with one of the most prevalent applications being search and rescue missions. These operations may be conducted in response to emergencies such as floods, fires, earthquakes, or in the case of a lost human in wilderness areas. In these scenarios, the operator may face considerable risks from the surrounding environment. During search and rescue efforts in complex settings like buildings or forests, the operator visual system can become overloaded. Therefore, it is critical to enhance the operator's access to visual information while minimizing the cognitive load on their visual processing capabilities. 

\begin{figure}[!ht]
    \centering
    \includegraphics[width=0.98\linewidth]{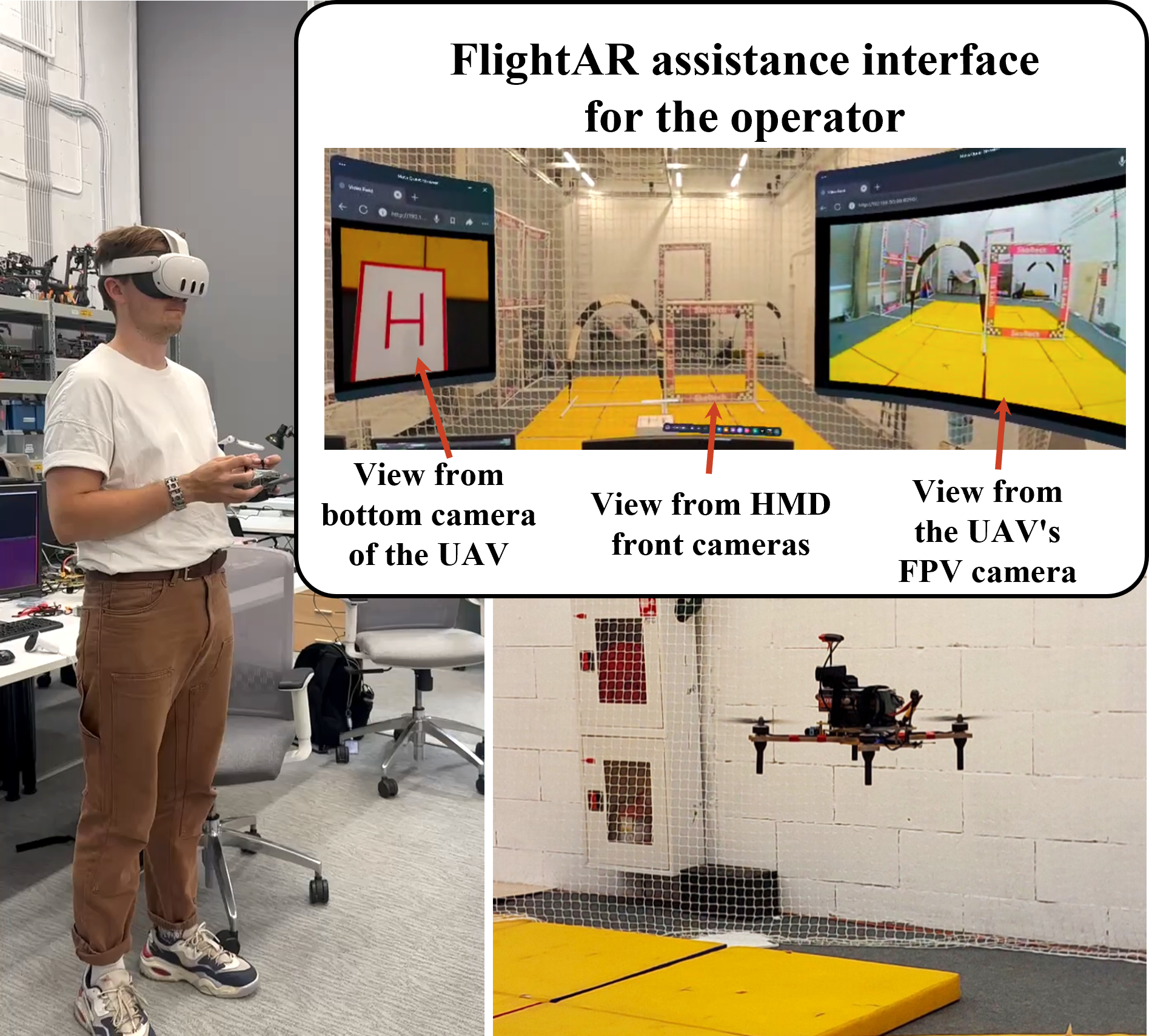}
    \caption{UAV operation using the developed FlightAR assistance interface.}
    \label{main_pic}
\end{figure}

Furthermore, UAVs frequently carry payloads such as cameras, LiDARs, magnetometers, and measurement instruments. Typically, these payloads require specific adjustments prior to flight, which may involve manual calibration or interfacing with the equipment through a computer or console. If the UAV operator is also responsible for maintaining the payload's functionality, they must concentrate on multiple critical elements of the system before takeoff. The implementation of augmented reality (AR) could reduce the cognitive load on the operator and eliminate the need to frequently remove the headset or shift their gaze.

The integration of AR and Artificial Intelligence (AI) technologies into flight assistive systems for UAV operators offers a promising approach to enhancing operational efficiency and safety of UAV navigation. With the rapid development of AI algorithms with low computational cost, UAV onboard computers can analyze the environment to detect objects, avoid obstacles, and optimize the flight path. Application of AR to overlay critical flight data and visualization of trajectory waypoints directly into the operator's field of view (FOV) could significantly improve real-time environmental awareness, enabling better decision-making.


This paper introduces FlightAR, a novel AR-based interface for human-drone interaction aiming to facilitate UAV control and enhance situational awareness through multiple views for the operator and real-time object detection. The proposed system utilizes a Meta Quest 3 head-mounted display (HMD) to provide the operator an intuitive visual interface represented by several video streams from the UAV cameras augmented with object detection with the ability to observe the real first-person view environment. The proposed system is aimed to greatly increase mission efficiency, safety, and operator's spatial awareness by making UAV operation more intuitive and responsive to the pilot's actions and environment change.


\section{Related Works}

\subsection{AR/VR-based Human-Drone Interaction Interfaces}

Currently, various research groups explore the application of virtual and augmented reality technologies to improve human-drone interaction (HDI). The application of AR and VR can enhance the user experience and operational efficiency. 

For example, Matrosov et al. \cite{matrosov2016lightair} developed a LightAir system that allows users to control and communicate with drones through foot gestures and images projected onto the ground. This system moves beyond traditional hand gestures, providing a more intuitive, natural, and safe interaction with drones, particularly for mid-sized quadcopters. A similar concept of projected-based visual interface was proposed by Cauchard et al. \cite{cauchard2019drone}. Drone.io is an HDI interface that comprises a projector-camera system embedded on a drone and a gesture recognition input system. Liu et al.\cite{liu2020augmented} designed an AR interface to enhance human-drone interaction during autonomous navigation tasks. The interface is deployed on a Microsoft HoloLens mixed reality glasses and enables users to interact with a 3D map reconstructed by the drone. 
Chen et al. \cite{chen2019pinpointfly} introduced PinpointFly, a mobile AR interface that allows pilots to directly interact with the drone position and orientation through visualization of the drone location and direction using a virtual cast shadow on the ground. This designed approach enables pilots to indicate or draw the desired flight path of the drone on the touchscreen, simplifying the control process and enhancing spatial perception. The interface is particularly effective in indoor and visual line-of-sight (VLOS) scenarios, where precise drone movements are essential. Mourtzis et al. \cite{mourtzis2024unmanned} proposed a method that integrates AR to facilitate remote path planning and control of UAVs in indoor environments, enhancing operational efficiency. The AR-based mobile application developed in this study uses Particle Swarm Optimization (PSO) to improve path planning flexibility compared to traditional algorithms such as A*, providing a more adaptable approach to mission design. Dorzhieva et al.\cite{dorzhieva2022dronearchery} presented  DroneARchery, a system that enhances human-drone interaction through the integration of AR and haptic feedback, supported by advanced deep reinforcement learning (DRL) algorithms. The system allows users to deploy a swarm of drones in an intuitive and immersive manner, simulating an archery shot. The use of a haptic interface provides tactile feedback that mimics the tension of a bowstring. The application of AR enables users to visualize the drone trajectories in real time, while the DRL algorithm manages collision avoidance among the drones. 

The integration of VR interfaces can facilitate the interaction of humans with autonomous robotic systems. Thus, Kalinov et al. \cite{kalinov2021warevr} proposed a VR-based interface for intuitive human interaction with the autonomous robotic system for stocktaking composed of a mobile platform and a UAV.
Yashin et al. \cite{yashin2019aerovr} presented  AeroVR, a VR-based teleoperation for aerial manipulation. The developed system allows operators to control a UAV equipped with a 4-DoF robotic arm in remote through a digital twin of the robotic system. The system provides real-time tactile feedback through a wearable control interface, enhancing precision and control for aerial manipulation tasks.

\subsection{Applications of Deep Learning in Human-Robot Interaction}
Deep learning technologies support human-robot interaction (HRI), guiding and improving robotic operations with real-time object detection. By leveraging advanced neural networks, robots can analyze vast amounts of data from sensors and cameras, allowing them to interpret complex environments and adapt their trajectories to avoid collisions. In addition, DL can help robots recognize when human operators are in close proximity, ensuring safe interactions in collaborative environments.

\begin{figure*}[!t]
\centering
\includegraphics[width=0.92\linewidth]{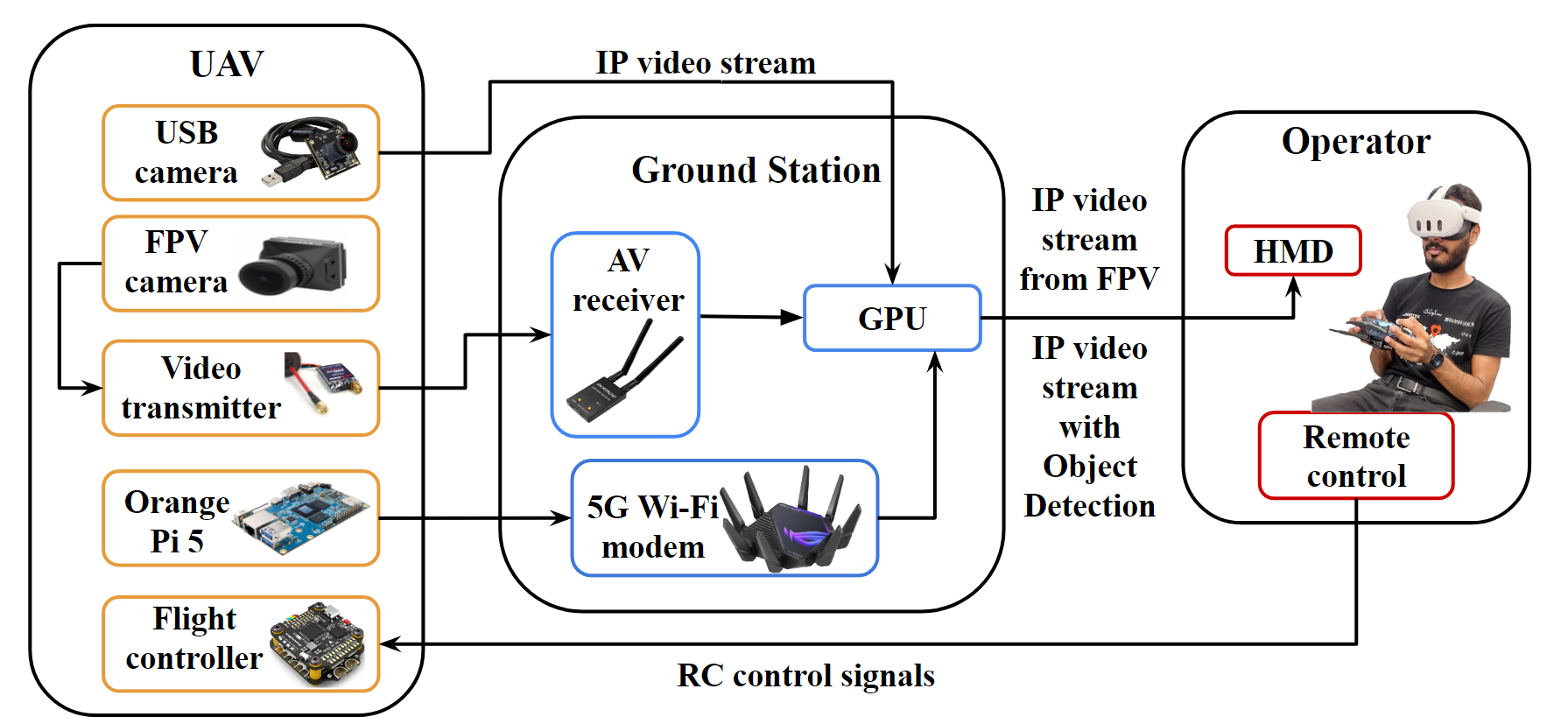} 
\caption{System architecture.}
\label{fig:architecture}
\end{figure*}

Lee et al. \cite{lee2020visual} introduced a telepresence system designed to enhance aerial manipulation tasks. The designed system leverages onboard visual and inertial sensors, object tracking algorithms, and pre-generated object databases to accurately recreate remote scenes for the operator. The proposed approach enhances the precision of tasks such as grasping, placing, force exertion, and peg-in-hole insertion by utilizing an extended marker tracking algorithm integrated with visual-inertial odometry.
Ponomareva et al. \cite{ponomareva2021grasplook} presented a VR-based telemanipulation system augmented with R-CNN-based detection and pose estimation of laboratory equipment objects for manipulation. The developed visual perception system detects the position of target objects, the coordinates of which are then transferred to the VR interface for rendering a 3D model of the recognized object. Similarly, Wonsick et al.\cite{Wonsick21} introduced an approach to recognize objects in remote environment using DL model and render high-fidelity 3D models of detected objects into VR interface. Nazarova et al. \cite{nazarova2021cobotar} presented an HRI system based on AR projector-based spatial display and Deep Neural Network (DNN)-based gesture recognition. The AR spatial display generates an interactive projected graphical user interface on a 
collaborative robot, which in combination with the gesture recognition system allows users to control the robot in an intuitive way.
Karmanova et al. \cite{karmanova2021swarmplay} proposed a human-swarm interactive system with reinforcement learning algorithm in the application of Tic-tac-toe game scenario to upgrade the level of user engagement and interactivity of HRI. Serpiva et al.\cite{serpiva2024omnirace} proposed OmniRace system that allows controlling racing drones using a 6-DoF hand pose estimation.

\subsection{Drone Object Detection Using Deep Learning Algorithms}

The development of object detection methods using DL algorithms introduced a significant advancement in aerial surveillance and monitoring. By leveraging convolutional neural networks (CNNs) and other DL techniques, these systems can accurately and efficiently identify objects in real-time, enhancing the capabilities of drones in various applications such as traffic monitoring, search and rescue operations, and environmental monitoring. Despite challenges such as computational demands and sensitivity issues, the integration of DL with drone technology promises a future of more responsive and intelligent aerial systems.
Thus, Kyrkou et al. \cite{kyrkou2018dronet} presented DroNet, a single-shot CNN for real-time vehicle detection using UAVs. The experimental results showed that DroNet achieved a 95\% detection accuracy with 5–10 frames per second (FPS) at on-board computers. In \cite{kalinov2020warevision}, a CNN-based real-time barcode detection and scanning system under various lightning conditions was proposed to optimize the UAV trajectory during autonomous warehouse stocktaking. 
Menshchikov et al. \cite{menshchikov2021real} proposed a system for real-time detection of hogweeds using UAV and a convolutional neural network running on a low-power embedded system. Zhao et al. \cite{zhao2023yolov7} introduced a model on YOLOv7 for detection of tiny-scale people or objects from maritime UAV images. The authors introduced a prediction head, implemented a parameter-free attention module, and utilized data augmentation techniques to achieve improved results.
Zhai et al. \cite{zhai2023yolo} proposed a model for UAV detection of tiny objects based on the optimized YOLOv8. The authors enhanced the model's performance by incorporating a high-resolution detection head, employing advanced feature extraction techniques, and utilizing an attention mechanism. 
In general, most autonomous flying robotic systems rely on lightweight detection models. In recent years, many of these systems have adopted transformer-based architectures, employing various approaches to enhance frames per second (FPS) performance.


\section{System Overview}
The UAV flight assistance system integrates both hardware and software to deliver an advanced real-time augmented reality experience for drone operation. The system layout is shown in Fig. \ref{fig:architecture}. The proposed system employs an Orange Pi 5 on-board microcomputer to receive and process the video feed from the UAV. The UAV is equipped with 2 cameras for capturing the real-time video streams. A GPU personal computer is utilized to process the obtained video streams and perform object detection. The YOLOV8n algorithm is applied for real-time object detection, specifically to identify and highlight persons within the video stream, and the  Meta Quest 3 HMD is used to display the AR assistive interface for the operator.  

\subsection{Hardware and Software Integration}
The hardware integration for the UAV flight assistance system focuses on synchronizing the operation of components on both the drone and the ground station. The drone is equipped with an FPV camera and a USB camera (downward-facing) to capture videos. The video stream from the FPV camera are transmitted through a video transmitter, while an on-board Orange Pi 5 microcomputer handles initial processing and converts the video from the USB camera into an IP stream. The flight controller, connected to the Orange Pi, manages flight control commands based on signals from the operator’s remote control (RC). This integration synchronizes video capture, real-time object detection, and AR display, advancing situational awareness and precision for drone piloting.

The video streams are transmitted to a ground station, where an Analog Video (AV) receiver and a 5G Wi-Fi modem receive the video streams from the FPV and USB cameras, respectively. The ground station includes a powerful GPU, responsible for running the YOLOV8n object detection algorithm in real-time. This software layer analyzes the video feeds in real-time, identifying and marking objects within the stream. The development environment comprised Python, with OpenCV and Flask used for video streaming, and YOLOV8n for executing the detection tasks.Once the object detection is processed, the GPU sends the augmented video streams to the visual interface of the operator.

The operator’s setup includes a Meta Quest 3 HMD, which displays the video streams in two browser windows -- one showing the USB camera feed and the other showing the FPV feed augmented with object detection overlays. In addition, the operator is provided with video feed from the front cameras of HMD that display the real environment. The operator also uses an RC to control the UAV, with real-time adjustments informed by the immersive AR interface.



\begin{figure}[h]
\centering
\includegraphics[width=0.88\linewidth]{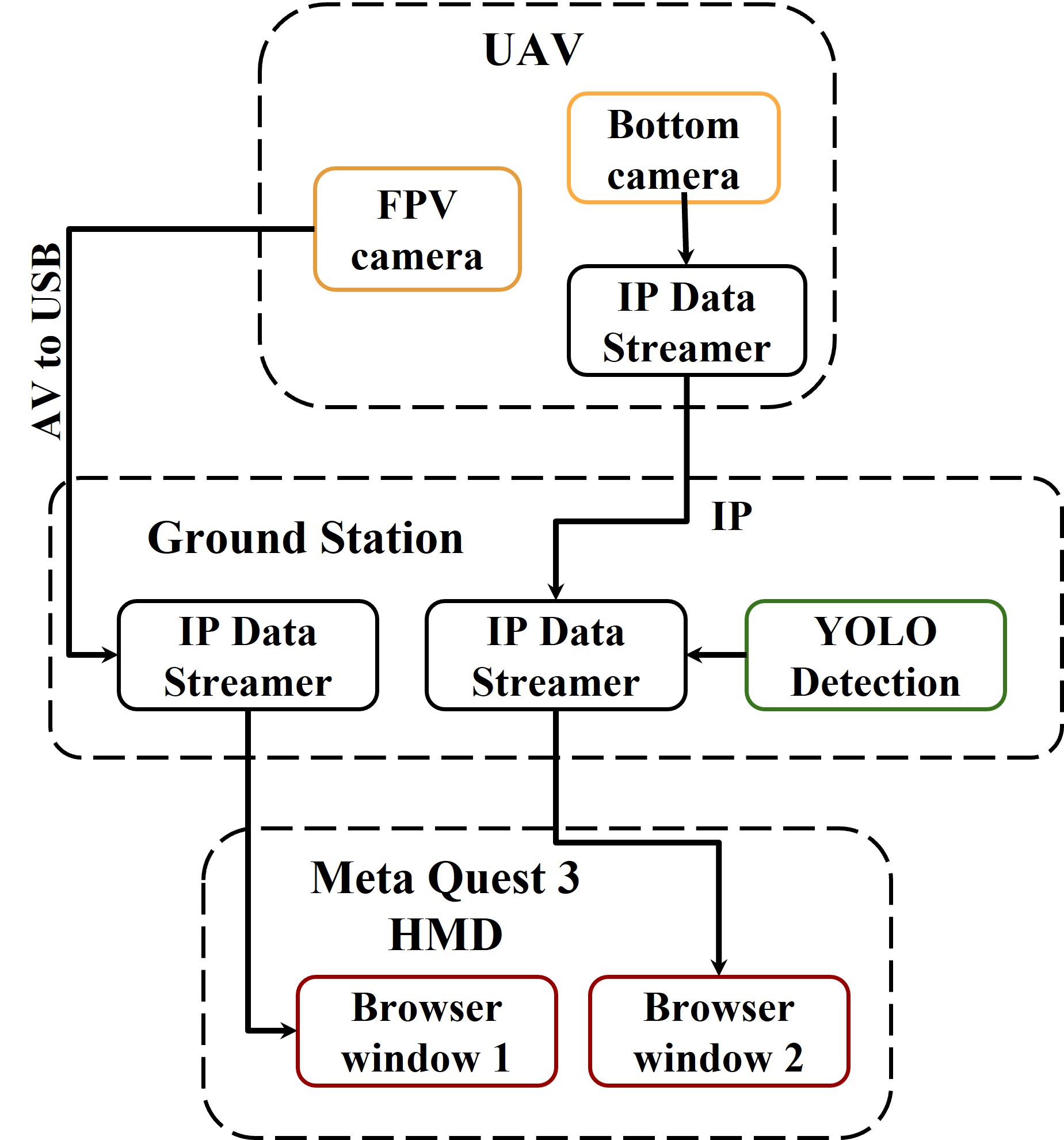} 
\caption{The pipeline of video transmission and processing.}
\label{software}
\end{figure}




\subsection{Visual Interface for the Operator}


Unlike conventional FPV goggles applied for drone piloting, the Meta Quest 3 HMD supports a passthrough operating mode which allows displaying real-time view of the user surrounding with 110$^{\circ}$ FOV using a set of front cameras. This feature allows integrating AR overlays in a transparent and seamless way with the pilot's visual channel. This helps the pilot to view augmented data without any obstruction, providing an immersive and interactive experience. The AR interface includes the processed video feeds from the UAV augmented with dynamic overlays of detected objects that are streamed to the Meta Quest 3 HMD through the browser application. The position and size of overlay windows could be easily adapted by the operator before the piloting. This setup significantly improves the pilot's ability to control the UAV by augmenting the visual channel with multiple views from the UAV that enhances both the efficiency and safety of UAV operations.

\subsection{Augmentation of Visual Channel with Object Detection using YOLOV8n Algorithm}
\begin{figure}[!h]
  \centering
  \subfigure[]{ \includegraphics[width=0.462\linewidth]{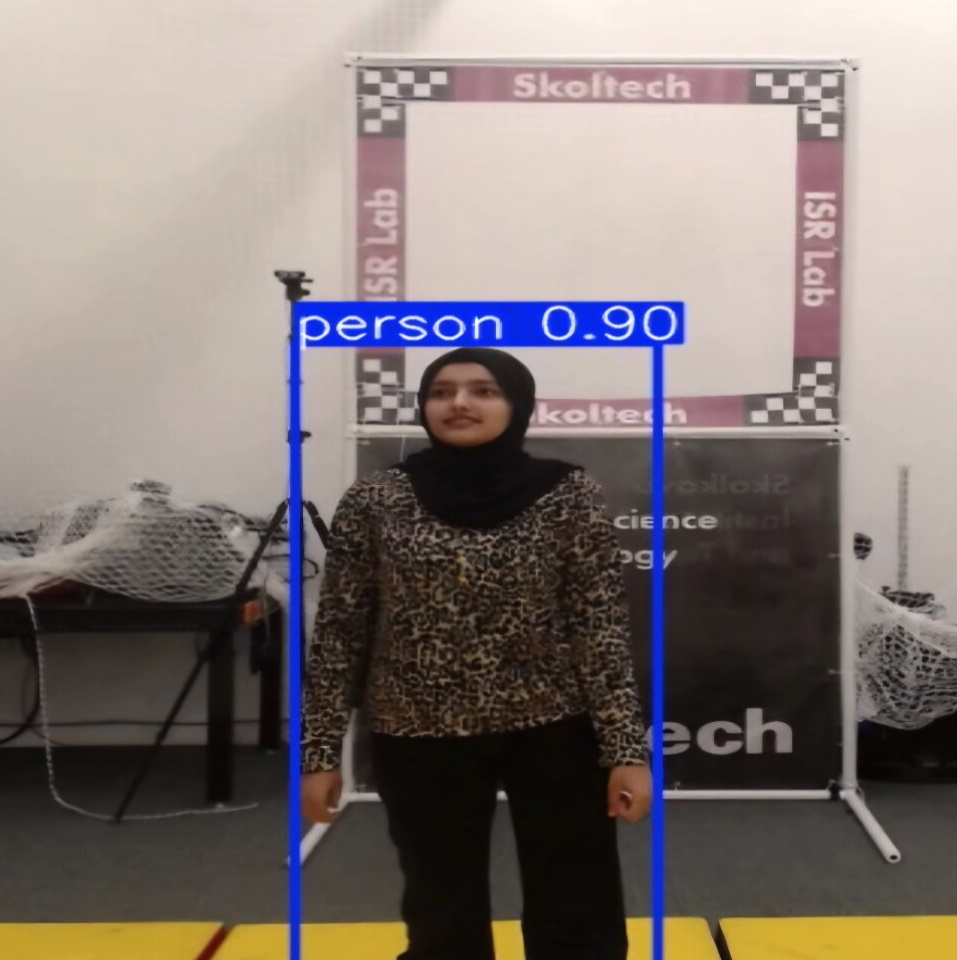}}
   \subfigure[]{\includegraphics[width=0.45\linewidth]{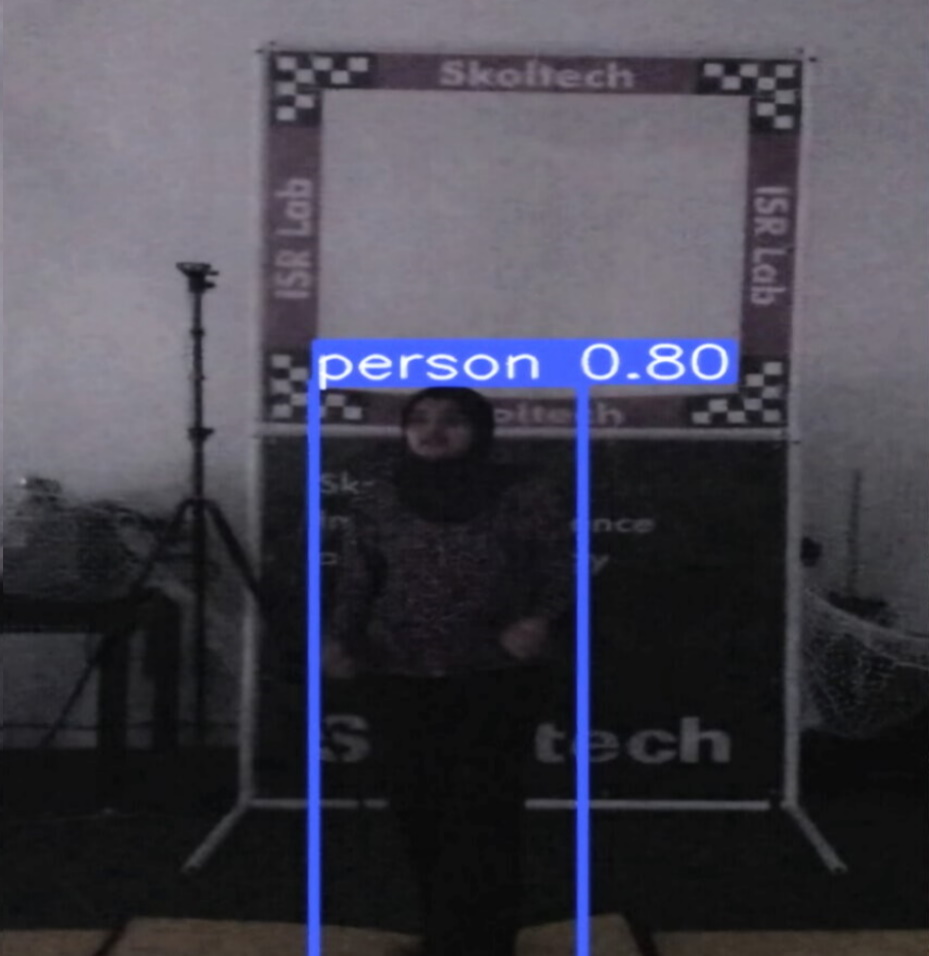}}
  \caption{The example of possible detection model usage in light (a) and dark conditions (b).}\label{fig:per_det}
\end{figure}

The object detection utilizes the YOLOV8n algorithm on the ground station computer to analyze the incoming video feed and detect persons within the frame. The algorithm dynamically draws bounding boxes around detected individuals in real-time, enhancing situational awareness for the pilot (Fig. \ref{fig:per_det}). By processing the entire image in a single forward pass of the network, YOLO achieves a high detection rate of 23.46 FPS. This makes it highly efficient for real-time applications where rapid and accurate object detection is crucial. YOLO's architecture, which combines CNNs with direct regression for bounding box prediction, sets it apart from traditional methods, allowing it to perform detection tasks with unprecedented rate and accuracy.

\subsection{The Limitations}
In the proposed setup, we utilized two methods to stream video data, namely Wi-Fi and an analog video signal. Currently, the system can operate at a distance of approximately 100 meters due to the limited range of Wi-Fi for video transmission over IP. However, this range can be extended up to 10 kilometers by employing a second analog video transmitter operating on a different frequency. For instance, one transmitter could use a frequency of 5.8 GHz while the other operates at 5.2 GHz. Additionally, a PC is required to receive the video and stream it over IP. If a detection model is needed, the setup should include a powerful GPU to ensure optimal performance.

\section{User study}
\subsection{Experimental Description}
The main goal of the developed system is to assist the pilot during the UAV operation with the simultaneous view from several UAV cameras, which will improve the control accuracy. 
For the experimental evaluation of efficiency of the proposed AR-assistant interface, we designed an experiment simulating the simplified conditions of a drone racing competition. 5 participants (all male) aged between 20 and 30 years old with experience in drone piloting, volunteered to participate in the experiment, giving their informed consent.

In this experiment, we compared two different visual interfaces to assess the participant's performance in drone piloting. The first was the commonly used FPV goggles, namely the Skydroid Cobra X model with 50$^{\circ}$ FOV. The second was the proposed FlightAR interface, designed to potentially improve usability and functionality.

\begin{figure}[t]
\centering
\includegraphics[width=0.95\linewidth]{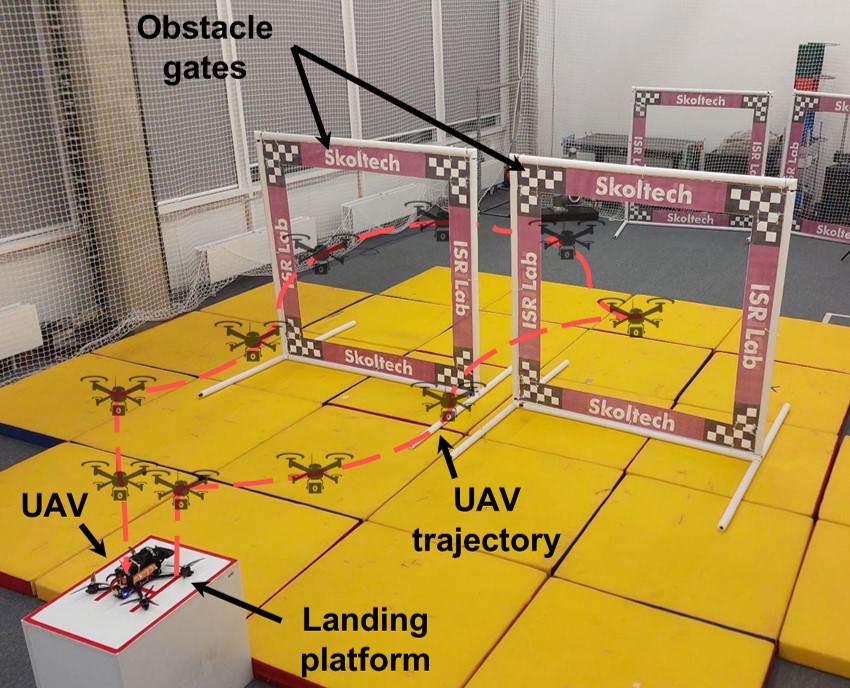} 
\caption{Experimental setup.}
\label{fig:exp_setup}
\end{figure}

\subsection{Experimental Procedure}
The task assigned to the participants involved two key phases: navigation and landing. Each pilot had to successfully fly through two designated drone gates, which tested their precision and control during the flight. Finally, they needed to land the drone accurately on a landing platform, showcasing their ability to maneuver the drone smoothly and with precision. Before the experiment, a short training session was conducted for each participant with instructions on the experimental procedure and familiarization with control interfaces. 
Each participant completed the flight task six times: three flights using the Cobra X goggles and three flights utilizing Meta Quest 3 HMD with assisted visual interface. For each participant, we measured the number of successful landings and participant workload during the task using NASA Task Load Index (NASA-TLX) questionnaire \cite{hart2006nasa}. In addition, we asked participants to evaluate their experience of using the designed FlightAR interface using User Experience Questionnaires (UEQ) \cite{ueq}.

\subsection{Experimental Results and Discussions}
\begin{figure}[h]
\centering
\includegraphics[width=0.92\linewidth]{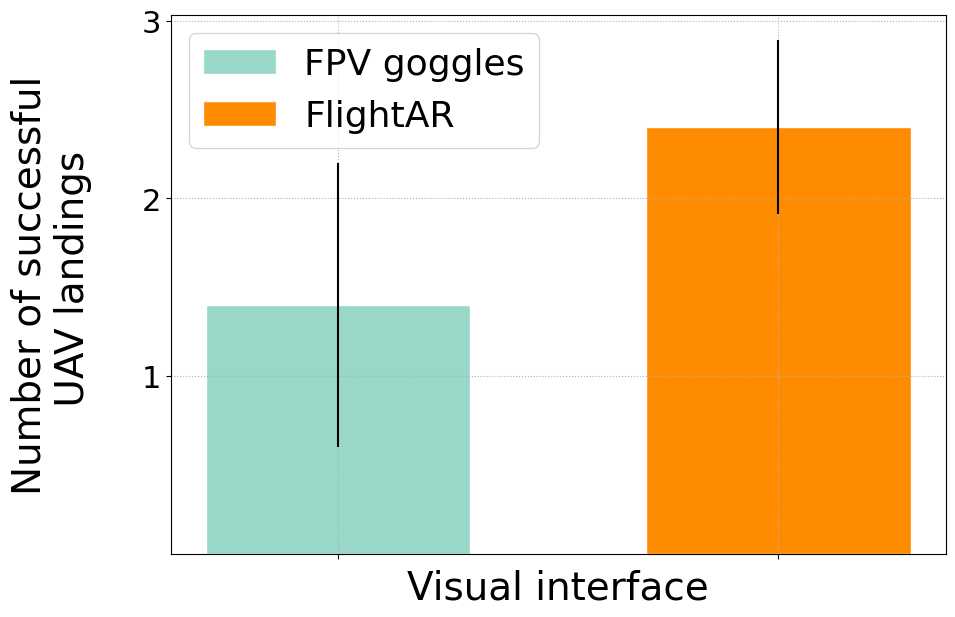} 
\caption{The statistics of successful UAV landing attempts.}
\label{fig:landing}
\end{figure}

Using the FlightAR interface, all participants successfully completed the UAV landing task in at least 2 out of 3 attempts (Fig. \ref{fig:landing}). While using FPV goggles, landing task performance statistics ranged from 0 (no successful landings) to 2 successful attempts. 

\begin{figure}
\centering
\includegraphics[width=0.92\linewidth]{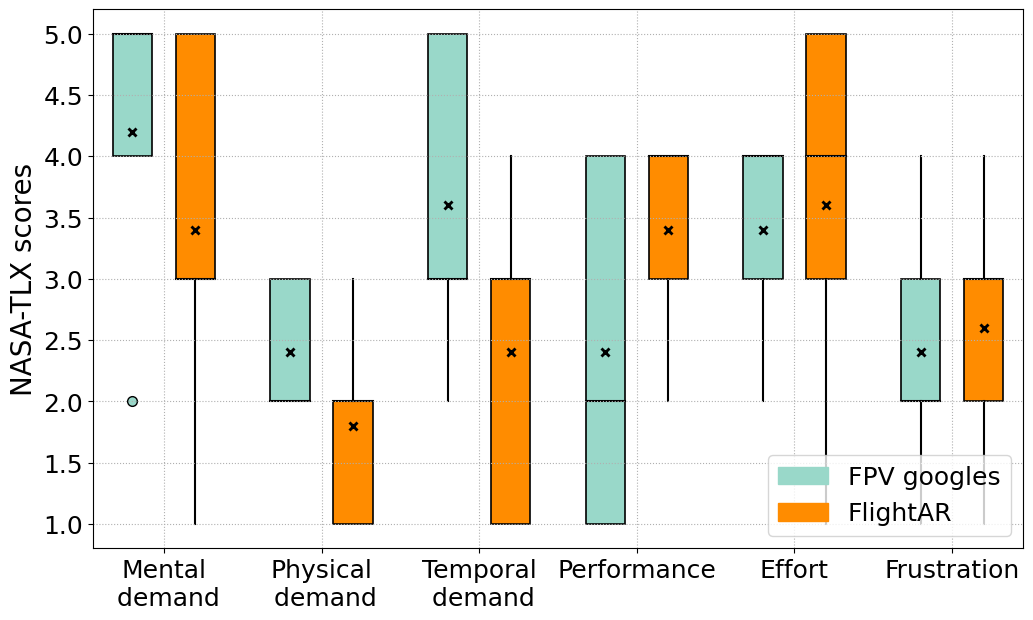}
\caption{NASA-TLX results in the form of a 5-point
Likert scale. Crosses mark mean values.} 
\label{fig:nasa}
\end{figure}

Evaluation of participants' workloads during the experiment using NASA-TLX is shown in Fig. \ref{fig:nasa}. Overall, the participants assessed low physical ($\mu=1.8$, $SD = 0.8$) and temporal ($\mu=2.4$, $SD = 1.2$) demands using the FlightAR interface, which were lower by 12\% and 24\%, respectively, than when using FPV goggles. Besides, participants rated their performance 20\% higher when using the FlighAR interface. At the same time, effort ($\mu=3.4$, $SD = 0.8$ and $\mu=3.6$, $SD = 1.5$ for FPV goggles and FlightAR, respectively) and frustration level ($\mu=2.4$, $SD = 1$ and $\mu=2.6$, $SD = 1$ for FPV goggles and FlightAR, respectively) were assessed similarly when using both interfaces. It can also be noted that using the proposed AR-assisted interface, the mental demand for the task decreased by 16\%.

\begin{figure}[h]
\centering
\includegraphics[width=0.92\linewidth]{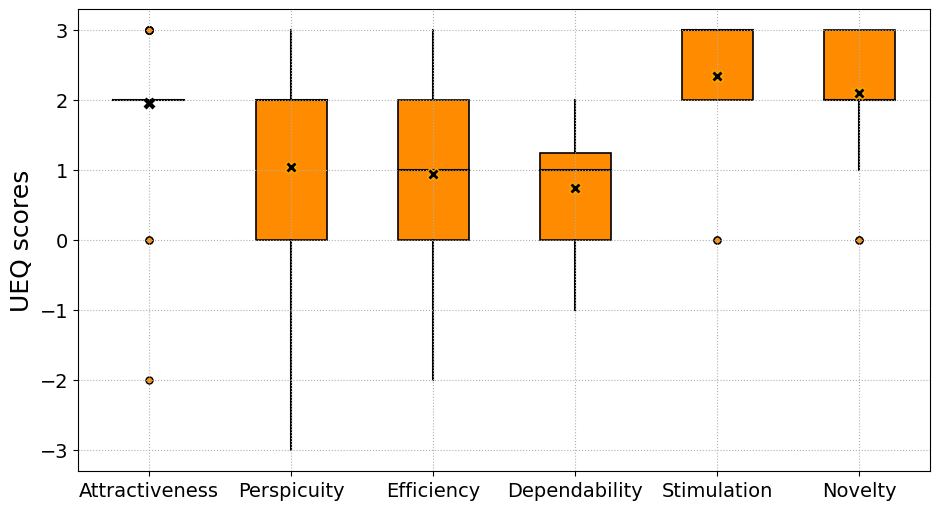} 
\caption{User Experience questionnaire results. Crosses mark average values.}
\label{fig:ueq}
\end{figure}

The results of the User Experience Questionnaire are shown in Fig. \ref{fig:ueq}. Overall, participants evaluated all the UEQ scales positively ($\mu>0.8$). It should be noted, that all participants highly rated their experience with FlightAR interface in terms of stimulation ($\mu=2.35$, $SD = 0.9$) and novelty ($\mu=2.1$, $SD = 0.9$). In addition, the attractiveness of the proposed FlightAR interface was also highlighted ($\mu=1.97$, $SD = 1$).

\section{Conclusion and Future Work}

In this work, we have proposed the AR-assistant interface to enhance UAV piloting performance by improving situational awareness and intuitive control. By integrating AR multiple video streams with direct visual channel of the pilot, pilots were able to gather crucial information from their environment more effectively than with traditional FPV goggles. The experimental results, supported by the NASA-TLX, revealed that the FlightAR interface reduced physical and mental demands by 12\% and 16\% compared to the FPV goggles. In addition, with FlightAR, pilots accessed higher task performance (by 20\%) compared to FPV goggles. A high user evaluation for stimulation ($\mu=2.35$, $SD = 0.9$), novelty ($\mu=2.1$, $SD = 0.9$), and attractiveness ($\mu=1.97$, $SD = 1$) indicate that the FlightAR system not only enhances technical performance but also offers a more engaging and satisfying user experience. These findings affirm the potential of the FlightAR interface to set new standards in UAV piloting, particularly in complex operational scenarios where situational awareness is crucial.

In future work, we plan to enhance the system for fast real-time detection in critical situations to expand its application in rescue operations. By upgrading the object detection algorithm to identify a broader range of items using a specialized rescue dataset, the system's performance can be made faster and more accurate. These advancements are expected to greatly improve the effectiveness and efficiency of drone-assisted rescue missions, making this technology a vital tool in emergency response efforts.

\section*{Acknowledgements} 
Research reported in this publication was financially supported by the RSF grant No. 24-41-02039.

\bibliographystyle{IEEEtran}
\bibliography{Ref}
\end{document}